\documentclass{article}

\usepackage{arxiv}

\usepackage[utf8]{inputenc} % allow utf-8 input
\usepackage[T1]{fontenc}    % use 8-bit T1 fonts
\usepackage{hyperref}       % hyperlinks
\usepackage{url}            % simple URL typesetting
\usepackage{booktabs}       % professional-quality tables
\usepackage{amsfonts}       % blackboard math symbols
\usepackage{nicefrac}       % compact symbols for 1/2, etc.
\usepackage{microtype}      % microtypography
\usepackage{lipsum}		% Can be removed after putting your text content
\usepackage{graphicx}
\usepackage{natbib}
\usepackage{doi}
\usepackage{amsmath}

\title{Self-supervised Pretraining of Cell Segmentation Models}

\author{ Kaden Stillwagon \textsuperscript{1*}, Alexandra Dunnum VandeLoo\textsuperscript{2, 3}, Benjamin Magondu \textsuperscript{3}, Craig R. Forest\textsuperscript{1,3,4} \\[0.5em]
\textsuperscript{1} School of Computer Science, Georgia Institute of Technology \\ 
\textsuperscript{2} School of Materials Science and Engineering, Georgia Institute of Technology \\ \textsuperscript{3} Department of Biomedical Engineering, Georgia Institute of Technology \\ \textsuperscript{4} School of Mechanical Engineering, Georgia Institute of Technology \\[0.5em] \texttt{*Corresponding author: kstillwagon26@gatech.edu}
}

\date{March 23, 2026}

\hypersetup{
pdftitle={Self-supervised Pretraining of Cell Segmentation Models},
pdfsubject={q-bio.NC, q-bio.QM},
pdfauthor={Kaden ~Stillwagon, Alexandra D.~VandeLoo},
pdfkeywords={Cell Segmentation, Self-supervised Learning, DINO},
}

\begin{document}
\maketitle

\begin{abstract}
	Instance segmentation enables the analysis of spatial and temporal properties of cells in microscopy images by identifying the pixels belonging to each cell. However, progress is constrained by the scarcity of high-quality labeled microscopy datasets. Many recent approaches address this challenge by initializing models with segmentation-pretrained weights from large-scale natural-image models such as Segment Anything Model (SAM). However, representations learned from natural images often encode objectness and texture priors that are poorly aligned with microscopy data, leading to degraded performance under domain shift.  We propose DINOCell, a self-supervised framework for cell instance segmentation that leverages representations from DINOv2 and adapts them to microscopy through continued self-supervised training on unlabeled cell images prior to supervised fine-tuning. On the LIVECell benchmark, DINOCell achieves a SEG score of 0.784, improving by 10.42\% over leading SAM-based models, and demonstrates strong zero-shot performance on three out-of-distribution microscopy datasets. These results highlight the benefits of domain-adapted self-supervised pretraining for robust cell segmentation.
\end{abstract}

% keywords can be removed

\section{Introduction}
Quantifying cell morphology is an integral part of an increasing number of research and clinical studies. In the process of growing living cells in a laboratory, called cell culture, cell quantification allows for unbiased monitoring through cell health prediction \cite{way_predicting_2021}, \cite{cole_light_2025} and senescence detection \cite{welter_machine_2024}, \cite{he_morphology-based_2024}, \cite{oja_automated_2018}, \cite{kamat_single-cell_2024}. Additionally, cell quantification has shown significant promise in culture analysis including being used to predict cell function \cite{vasilevich_correlation_2020} and gene expression \cite{nassiri_systematic_2018} as well as distinguish between cancerous and non-cancerous cells \cite{wu_single-cell_2020}, \cite{mousavikhamene_morphological_2021}. The technique that allows biologists to quantify cell morphology is instance segmentation, in which all pixels for each cell are identified. 
Early approaches to segmentation involved hand-drawing instance masks over each cell. This task is not only time-consuming as a single image can contain hundreds of cells but also requires domain expertise to be able to correctly distinguish cells that are tightly packed or have unclear boundaries. In addition, disagreement between annotators is common and can make it difficult to compare across studies. Any quantitative analysis requiring costly hand-drawn segmentation of large datasets of images was simply infeasible. As a result, automated tools were developed to increase annotation speed and reduce the workload required for morphological analysis. Well-known biology softwares such as ImageJ \cite{schneider_nih_2012} developed semi-automated cell segmentation through intensity-based thresholding techniques for binary segmentation with watershed \cite{roerdink_watershed_2000} for instance separation. These methods came with severe limitations including problems with clumping cells together, misidentifying boundaries, and failure to recognize low-intensity cell parts. 

Modern deep learning approaches have since become the gold-standard for cell segmentation, and these advances are described in the Related Works section. The latest models use pretrained weights from segmentation-pretrained foundation models like the Segment Anything Model (SAM) \cite{kirillov_segment_2023}. While these methods greatly improve model performance, they come with their own inherent limitations: domain shift between pretraining and fine tuning data sets. In the cell domain, where performance ultimately matters, edges of objects are far less distinct than in pretraining data sets; there are only sparse semantic boundaries, and object shapes are highly varied and unusual. As a result, these models struggle to generalize to microscopy images outside of their training distribution. 

Therefore, we propose a new approach: adapting a DINOv2 \cite{oquab_dinov2_2024} image encoder to the domain of microscopy images as an initialization strategy for cell segmentation. In essence, while SAM is able to learn a strong task-specific prior in its pretraining, DINO can learn a strong domain prior through self-supervised domain adaptation to microscopy images. Our hypothesis is: due to the extreme domain shift between real-world images and microscope images, learning domain-appropriate visual representations is a more difficult and limiting problem than learning the segmentation mapping itself. Therefore, initialization weights with a domain prior should be more effective than initialization weights with a task prior.

\section{Related Works}
\subsection{Deep Learning for Cell Segmentation}
Deep learning has become the dominant approach for instance segmentation in microscopy images. The introduction of the U-Net \cite{ronneberger_u-net_2015} architecture demonstrated that encoder–decoder networks trained on biomedical images can significantly outperform classical thresholding-based techniques. Subsequent works \cite{falk_u-net_2019}, \cite{stringer_cellpose_2021} extended this design with improved loss formulations and architectural refinements for better boundary delineation and instance separation.

Many modern cell segmentation models reformulate the task by predicting intermediate objective representations rather than directly regressing instance masks. For example, flow-based approaches such as Cellpose \cite{stringer_cellpose_2021} predict pixel-wise vector fields that point toward cell centers and use gradient-tracking algorithms to reconstruct instances. Related methods including Omnipose \cite{cutler_omnipose_2022} extend this idea by refining objective definitions to improve performance on elongated or densely packed cells. These proxy-based strategies improve separation between touching cells but remain heavily dependent on the quantity and quality of annotated training data. Despite these improvements, limited labeled datasets and annotation inconsistencies continue to restrict generalization across diverse imaging conditions.

\subsection{Foundation Models for Cell Segmentation}
The emergence of large-scale vision foundation models has reshaped representation learning for segmentation tasks. Models such as the Segment Anything Model (SAM) \cite{kirillov_segment_2023} demonstrate that training on large-scale image corpora can produce transferable visual features useful for downstream segmentation. Several recent cell segmentation methods leverage pretrained SAM encoders as initialization backbones and replace the prompt-based decoder with task-specific prediction heads \cite{ma_segment_2024}, \cite{vandeloo_samcell_2025}, \cite{marks_cellsam_2025}, \cite{pachitariu_cellpose-sam_2025}. These approaches benefit from strong pretrained representations but often require significant architectural adaptation to handle dense microscopy images containing many objects per frame. Additionally, the domain gap between natural images and microscopy imagery limits the effectiveness of task priors learned from large-scale segmentation supervision. While foundation model initialization improves fine-tuning performance, their representations are primarily optimized for natural-image object segmentation. This mismatch motivates exploration of alternative pretraining strategies that better align with microscopy data distributions.

\subsection{Self-Supervised Representation Learning}
Self-supervised learning provides a mechanism to learn visual representations from unlabeled data by constructing proxy tasks directly from images. Contrastive learning methods and masked autoencoding strategies have demonstrated strong transfer performance across vision tasks without requiring manual annotations \cite{he_masked_2021}, \cite{chen_simple_2020}. Self-distillation approaches, such as DINO \cite{caron_emerging_2021} and its improved variant DINOv2 \cite{oquab_dinov2_2024}, learn image representations through a teacher–student framework that encourages consistency across augmented views. These methods have been shown to produce spatially coherent embeddings that capture objectness, semantic similarity, and part-level structure. Importantly, DINOv2 trained on large-scale curated datasets has demonstrated strong transfer performance across diverse downstream tasks, including dense prediction problems \cite{zhuo_uino-fss_2025}, \cite{baharoon_evaluating_2024}, \cite{cui_surgical-dino_2024}.

Despite their success in general computer vision, relatively little work has explored adapting large self-supervised vision transformers specifically to microscopy imagery prior to segmentation training. Domain-adaptive self-supervised training offers a promising approach to align pretrained representations with the structural characteristics of biological images. In this work, we leverage self-supervised representation learning as a domain adaptation mechanism to construct microscopy-aware visual features before supervised segmentation fine-tuning.

\section{Method}
\subsection{Data}
We collate two non-overlapping curated datasets in this work: 1) an unlabeled set of images for self-supervised domain adaptation and 2) a labeled dataset of cell images with hand-drawn segmentations for fully supervised fine-tuning. All datasets used are publicly available. 

\subsubsection{Domain Adaptation Dataset}
The dataset we curate for domain adaptation of pretrained DINOv2 weights consists of ~130,000 unlabeled cell images. We collect these images from subsets of 63 independent sources released in public biodata banks including the Broad Institute Cellpainting Gallery (CPG) \cite{weisbart_cell_2024}, Broad Bioimage Benchmark Collection (BBBC) \cite{ljosa_annotated_2012}, and Image Data Resource (IDR) \cite{williams_image_2017}. Together, this dataset encompasses over 50 cell types across four image modalities: phase-contrast (PC), brightfield (BF), differential interference contrast (DIC), and fluorescence (Fl). We take a 1024x1024 crop from each of these images and bias the crop to areas of high cell density by convolving a Laplacian filter over the image and selecting the patch with the highest variance. This minimizes the number of training images containing no or very few cells as can happen in sparse cultures. For memory efficiency, we compress all images to grayscale. This presents a particular challenge for fluorescence images where channels don’t represent colors, but the intensity of light emitted by different fluorophores after they have been excited. To account for the various ways biologists analyse these images, we separate the channels into independent images and also include a combined image. This dataset is highly varied and encapsulates a much larger portion of cell diversity than most labeled datasets.

\subsubsection{Fine-tuning Dataset}
We finetune DINOcell on 15 publicly available annotated datasets containing a combined 9,042 images (after filtering). All images in the Fine-tuning Dataset are live, cultured cells and include three modalities: PC, BF, and DIC. Additionally, for all images in the dataset, the ground truth data labels entire cells (rather than labeling only the nucleus as is done within other datasets).

In our investigation of these datasets we find a large amount of unclean data including corrupted images, image artifacts, missing labels, and non-biological data. We handle this polluted data in two ways: first pass filtering and ignore masking. Our initial filter removes erroneous examples (e.g. 0-variance images, fully misaligned segmentations, non-biological data) through visual inspection of images and their overlaid segmentation masks. We utilize ignore masking to mitigate the detrimental effects of examples that are missing some segmentations by down-weighting training loss for all unlabeled pixels. See the "Training" section for further discussion of this functionality. We divide the remaining data into train, validation, and test sets. For datasets without predefined splits, an  80/10/10 train/validation/test split was used.

The datasets are named based on their origins, as described in the “Supplementary Materials.”  Specifically, there are tens to thousands of images in each of the following: BBBC030 \cite{koos_dic_2016}, BCCD \cite{dpto_blood_2023}, Cellpoose cyto (live) \cite{stringer_cellpose_2021}, DeepBacs E.Coli (BF) \cite{spahn_deepbacs_ecoli_bf_oct_2021}, DeepBacs E.Coli (Phase) \cite{spahn_deepbacs_2022}, DeepBacs Staph Aureus \cite{pereira_deepbacs_2021}, NeurIPS 2022 Challenge \cite{ma_neurips_2024}, Omnipose Bact Phase \cite{cutler_omnipose_2022}, Omnipose Worm (high res) \cite{cutler_omnipose_2022}, Omnipose Worm \cite{cutler_omnipose_2022}, YeaZ \cite{dietler_convolutional_2020}, idr0095 \cite{williams_image_2017}, LIVECell \cite{edlund_livecelllarge-scale_2021}, NTERA-2 \cite{zhang_morphoseg_2025}, and TYC \cite{vijayan_deep_2024}. These datasets are not only used for training and validation, but also used to compare the relative performance (using the Maximum Matching Accuracy (MMA) metric as described below) of DINOCell between them [see “Performance on Individual Datasets” section].

\subsubsection{Datasets for Evaluation and Comparison}
When evaluating DINOCell models against each other, we measure performance on the full test set from our Fine-tuning Dataset. This set contains 2101 images from all 15 sources. In comparing DINOCell to other models, we evaluate all models on the LIVECell \cite{edlund_livecelllarge-scale_2021} test set for in-distribution performance and the three out-of-distribution datasets (PBL-HEK, PBL-N2A \cite{vandeloo_samcell_2025}, and Glioma C6 \cite{malashin_glioma_2025}) for zero-shot performance.  

 LIVECell is commonly used as an evaluation benchmark to compare cell segmentation models \cite{edlund_livecelllarge-scale_2021}, \cite{reith_selfadapt_2025}, \cite{wang_systematic_2024}, \cite{khalid_point2mask_2022}, \cite{das_high-throughput_2025}. The LIVECell dataset contains 5,239 labeled phase-contrast images including eight distinct cell types with highly variable morphology and culture densities. It contains a predefined train/test split and was included in the training of DINOCell and all competitor models.  
 
 We created the PBL datasets for  zero-shot evaluation of cell segmentation models. The two datasets each contain five Phase-Contrast images of HEK and N2A cells respectively, imaged with varying magnification and contrast and at differing confluency levels. These images were hand-annotated by a biologist with extensive cell culture experience. The specific microscope conditions and cell types represented by these images are not found in the training sets of DINOCell or any other models in consideration.  
 
 The Glioma C6-gen dataset contains 30 images of 3900 individual Glioma cells and was built specifically to test the generalizability of trained segmentation models. The dataset includes two cell types (Type A and Type B cells) with distinct morphology including spheroid, spindle-shaped, elongated, and disk-like cells. In addition, these images were acquired under varying imaging and seeding conditions to further test model robustness. 

\subsection{Architecture}
\subsubsection{Pre-Processing}
The first step in our model pipeline is to prepare input images for evaluation through preprocessing. We apply a combination of Contrast Limited Adaptive Histogram Equalization (CLAHE) \cite{zuiderveld_contrast_1994} and normalization to zero-mean and unit variance. Together, these help to sharpen cell edges, remove brightness artifacts, and stabilize model convergence. Images smaller than the crop size of 512x512 are resized so that the shortest side of the image is 512 in length and the aspect ratio is maintained. Lastly, we take a 512x512 crop of the image and upsample it bicubically to size 896x896. 

\subsubsection{Model Design}
The DINOCell architecture processes images in two stages: encoding and decoding. Encoding is performed by a ViT-B image encoder that has been initialized with DINOv2 pretrained weights and domain adapted on our Domain Adaptation Dataset. We reduce the default DINOv2 patch size of 14x14 to 8x8 requiring us to resize the position embeddings and projection layer weights. The encoder produces 112x112 patch embeddings with an embedding dimension of 768.  

\begin{figure}[htbp]
\centering
\includegraphics[width=1\textwidth]{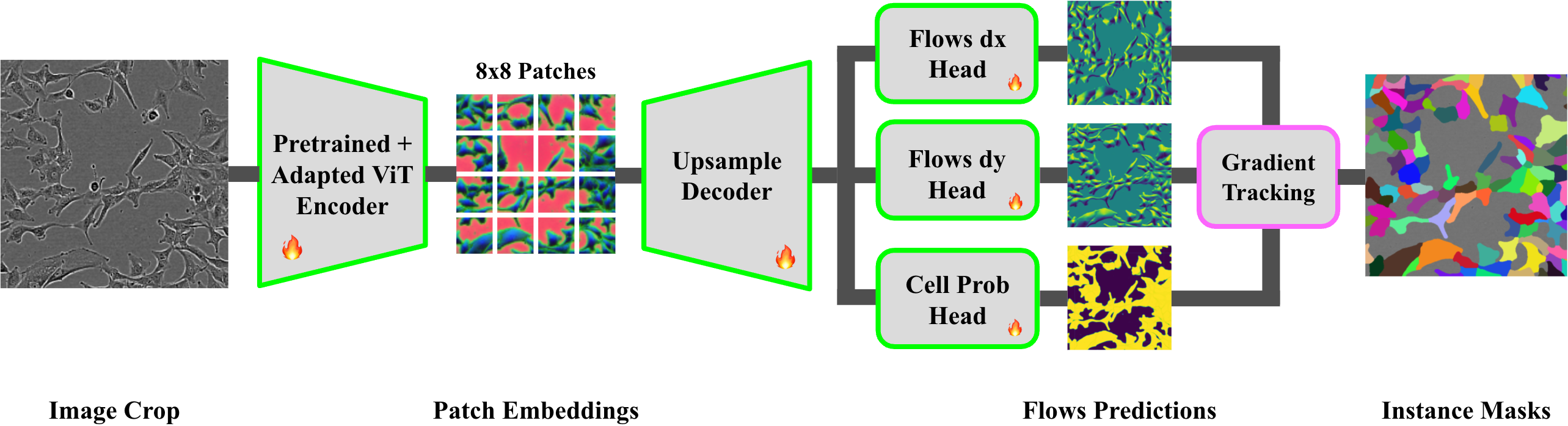}
\caption{\textbf{DINOCell Model Architecture.} Schematic of the DINOCell model architecture containing a DINOv2-pretrained \cite{oquab_dinov2_2024} ViT-B \cite{dosovitskiy_image_2021} encoder that has been domain-adapted on microscopy images, a lightweight upsampling decoder, and convolutional objective prediction heads. The model outputs are combined and decoded using gradient-tracking to produce instance segmentation masks.}
\label{fig:Architecture}
\end{figure}

These embeddings are then fed into a lightweight upsampling decoder. This decoder is composed of two transposed convolution blocks and one upsampling block that progressively increase the spatial dimension of the embeddings to their original size of 512x512 while simultaneously shrinking the embedding dimension to a 1/8th of its initial depth. Small prediction heads consisting of three convolutional layers convert these decoded embeddings into objective maps (e.g. flows\_dx, flows\_dy, and cell probability for Cellpose-style flows). Figure \ref{fig:Architecture} provides a full schematic of the DINOCell architecture.

\subsubsection{Model Objective and Post-Processing}
Instead of directly outputting segmentation masks like U-Net models do for semantic segmentation or SAM does with promptable segmentation, we take inspiration from works including SAMCell and Cellpose and predict objective proxies from which segmentation masks can be mathematically derived. These objective proxies have been empirically shown to improve separation of cells with unclear boundaries. Distance map and flows objectives are found most prominently in literature \cite{vandeloo_samcell_2025}, \cite{pachitariu_cellpose-sam_2025}, \cite{kucharski_cnn-watershed_2021}. Each of these objectives involve two associated functions: one to convert ground truth masks into objective targets and another to convert model objective predictions into instance segmentation masks. We find that Cellpose’s flows achieves the highest test and zero-shot accuracy when used as the objective proxy for DINOCell.  
The flows objective is a vector flow field representation (see Figure \ref{fig:Architecture}). To transform a manually annotated mask into a flow, a simulated diffusion process starting at the center of the mask is used to derive spatial gradients that point towards this center. Predicted gradient fields are then used to construct a dynamical system where each pixel follows the gradients to their eventual fixed point, enabling association between pixels that end at the same final point. 

\subsubsection{Sliding Window Pipeline}
At inference time, DINOCell uses a sliding window approach to process images that are larger than the 512x512 crop size. Specifically, the input image is sliced into 512x512 crops and our finetuned model generates objective predictions on each crop independently. We use an overlap size of 128 on all sides of the crop and average the overlapping objective predictions to mitigate the effects of inference on partially observed cells at crop boundaries. Full-image segmentation masks are then decoded from the recombined objective image.  

\subsection{Training Methodology}
DINOCell is trained in three phases: 1) self-supervised pretraining with DINOv2, 2) domain adaptation to microscope images through continued DINO training, and 3) fine-tuning on human-segmented cell images. We seek to retain and strengthen the strong inductive biases learned during pretraining so that the final model may more effectively generalize to unseen images.

\subsubsection{Self-supervised DINOv2 Pretraining}
Self-supervised training aims to learn transferable representations from unlabeled data by constructing artificial supervision signals from the data itself. DINOv2 learns visual representations through self-distillation within a student-teacher framework. The model consists of a student ViT network and a momentum-updated teacher ViT network each with independent weights. For each training image, global and local crops are generated and uniquely augmented. The student then processes all crops, while the teacher processes only the global crops. Both networks project their embeddings to a vector of “prototype scores” that is normalized with temperature softmax. The similarity of these vectors is measured through a cross entropy loss and the student weights are optimized to match the teacher’s output distribution. The teacher’s parameters are updated as an exponential moving average of the student network’s weights. Importantly, DINOv2 adds an $\textbf{i}$mage $\textbf{B}$ERT pre-training with $\textbf{O}$nline $\textbf{T}$okenizer (iBOT) \cite{zhou_ibot_2022} objective to the training formula where randomly masked student patch tokens are trained to predict the corresponding unmasked teacher token representations. Output centering, sharpening, and strong data augmentation are used to prevent representation collapse and stabilize optimization. Together, these objectives produce semantically structured and spatially coherent features that transfer effectively to downstream dense prediction tasks (e.g. segmentation). The DINOCell image encoder is initialized with the pretrained weights from the ViT-B DINOv2 model. 

\subsubsection{Domain Adaptation}
Domain adaptation is the process of adjusting a model that was trained on a particular domain so that its representations perform well on a target domain by minimizing the gap between their data distributions. In this work, we conduct domain adaptation on a pretrained DINOv2 model through continued DINO training on a dataset of images from our target domain: microscopy. Our objective is to reduce the distributional mismatch between natural-image pretraining data and microscopy images, thereby aligning learned feature representations with domain-specific structural patterns while preserving generalizable visual priors. We resume DINO-style training on our dataset of unlabeled 1024x1024 cell images using an AdamW \cite{loshchilov_decoupled_2019} optimizer with a base learning rate of 2e-4, and a drop rate of 0.2. Local crops were resized to 210, global crops to 508, and 8192 prototypes were used by both DINO and iBOT heads. Training was split across eight L40S GPUs each with a batch size of 16. We train for 100 epochs where an epoch is defined as a random sample of 1000 images per GPU, for a total of 8000 images per epoch, requiring approximately 100 hours for domain adaptation. All other hyperparameters were unchanged from DINOv2 settings. 

\subsubsection{Segmentation Fine-tuning}
\begin{figure}[htbp]
\centering
\includegraphics[width=1\textwidth]{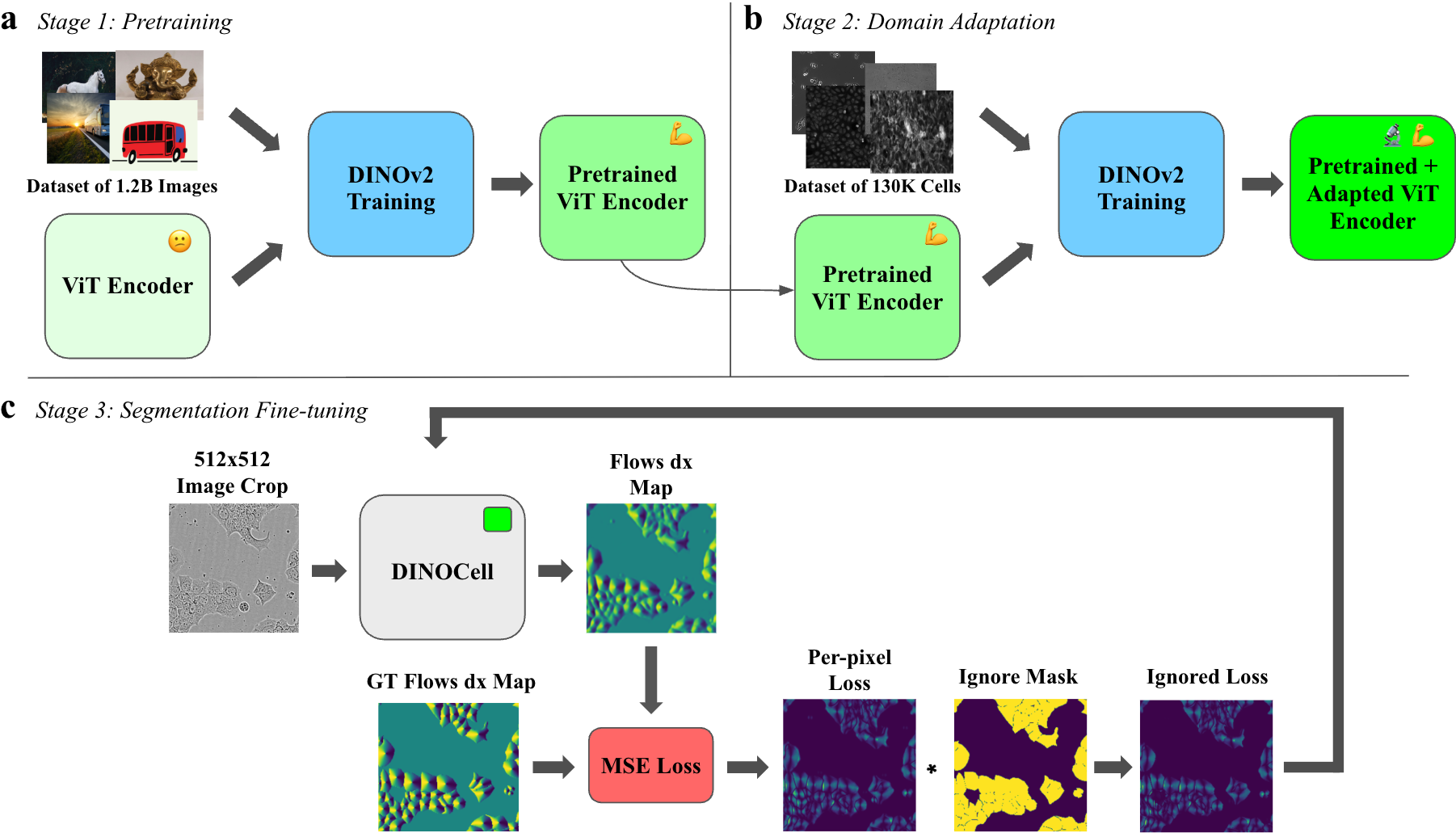}
\caption{\textbf{Three Stage Training Process of the DINOCell Model.}\textbf{(a)} Stage 1: A ViT-B \cite{dosovitskiy_image_2021} image encoder is pretrained on 1.2 billion images using DINOv2 \cite{oquab_dinov2_2024}, producing strong object- and part-aware visual embeddings. DINOCell is initialized with these pretrained weights. \textbf{(b)} Stage 2 (Domain Adaptation):  The encoder is further trained with DINOv2 on a curated dataset of 130k unlabeled microscopy images to adapt the representation to cell morphology. \textbf{(c)} Stage 3 (Fine-tuning):  The full DINOCell model is trained for cell segmentation with the encoder unfrozen. The model predicts flow fields (dx, dy) \cite{stringer_cellpose_2021} and cell probability from an input image. A mean squared error loss is computed between predictions and ground truth, reweighted using an ignore mask, and backpropagated to update model parameters.
}
\label{fig:Training_Stages}
\end{figure}

Fine-tuning is the final stage in the training process where the model attempts to learn the target task on the desired domain using fully supervised data. For DINOCell, this means training our full model architecture to produce segmentation objective proxies (i.e. flows) from which we can derive instance segmentations. Our training utilizes an AdamW optimizer with a weight decay of 1e-4 and a learning rate of 2e-5 that undergoes a warmup during the first 8 epochs and decays from there. Empirical testing reveals that it is best to unfreeze the pretrained image encoder and use a consistent learning rate throughout the model. Loss is computed per-pixel using mean-squared-error on each objective output (i.e. flows\_dx, flows\_dy, and cell probability for the flows objective) and then summed. Ignore masking is used to down-weight loss in unlabeled areas and requires loss to be renormalized for all non-ignored pixels. We train DINOCell for 40 epochs on an H200 GPU with a batch size of 8, requiring about 60 hours of training time. The full training process is outlined in Figure \ref{fig:Training_Stages}.

\subsubsection{Data Augmentation}
During DINO pretraining and domain adaptation, data augmentation is an integral part of the training loop for increased robustness and semantic understanding in feature representations. We employ the following augmentation: random (1) cropping and resizing between 5\% and 32\% for local crops and between 32\% and 100\% for global crops, (2) horizontal mirroring (p=0.5), (3) intensity jittering with brightness and contrast factors of 0.4, (4) gaussian blurring (p=0.01), and  (5) solarization with a threshold of 128 (p=0.1). During fine tuning we use very light intensity-based augmentations including random (1) brightness adjustment between 95\% and 105\% of original and (2) inversion of the image.  More intense augmentations were attempted for fine tuning but found to decrease both validation and zero-shot performance. Additionally, the directional nature of the flows objective prevents geometric augmentations without recalculating the ground truth values, which would slow training speed substantially.  

\subsubsection{Ignore Masking}
Many publicly available cell segmentation datasets have a common problem of missing segmentations in areas where there is clearly a cell. This can often occur when annotators are unsure of the exact way to segment a group of cells and elect to leave the area blank as opposed to making potentially incorrect annotations. These missing segmentations poison datasets and cause training to be unstable. When a model makes a correct prediction in an area with missing segmentations, this generates an unintended false positive error that biases the model’s subsequent predictions. This makes it difficult for the model to learn a generalized understanding of the task and forces the model to optimize to inaccurate data, causing overfitting. 

We address this problem with a technique called ignore masking. In ignore masking, pixels that are not associated with any ground truth label are downweighted to reduce the effect they have on the gradient update. We accomplish this by creating a binary mask that identifies all unlabeled pixels and multiplying the mask by a weight between 0 and 1 (we use 0.05). During training, the per-pixel loss image is multiplied by this weighted mask and renormalized by dividing the result by the sum of the total image weight where non-ignored pixels have a weight of 1.  The equation below \ref{eq:ignore_mask_loss} describes this calculation where L is the per-pixel loss and M is the mask where labeled pixels are 1 and unlabeled pixels are the ignore mask weight (0.05).

\begin{equation}
  L_{\text{ignored}} = \frac{L * M}{sum(M)}
  \label{eq:ignore_mask_loss}
\end{equation}

The effect of ignore masking is illustrated in Figure \ref{fig:Training_Stages}. We find that with proper tuning of the ignore mask weight, this technique can almost completely eliminate the problem of overfitting to missing ground truth labels. However, this method does come with the drawback that false positive predictions are penalized significantly less. We observe that this side effect is accompanied by increased mask bloating and oversegmentation early in training.  With sufficient training length and tuning of the ignore mask weight, however, these issues are mostly overcome by the end of training.

\subsection{Evaluation}
\subsubsection{Evaluation Metrics}
We use three metrics to evaluate inter and intra model comparisons: two from the Cell Tracking Challenge (SEG and DET) and a metric we developed in the course of this work, Maximum Matching Accuracy (MMA). Each of these metrics are measured on an interval between 0 and 1 with higher performance indicated by values closer to 1. 

\textit{SEG}: Segmentation Accuracy Measure (SEG) \cite{maska_benchmark_2014} measures the quality of instance segmentation by computing the average Intersection-over-Union (IoU) between matched ground-truth and predicted instances. Ground-truth and predicted masks are matched under a one-to-one constraint using an IoU threshold of 0.5 to determine valid correspondences. SEG is defined as the average IoU over all matched instances, with unmatched ground-truth instances not contributing to the score. This metric emphasizes boundary accuracy and penalizes both under- and over-segmentation. A formula \ref{eq:seg_metric} for calculating SEG is shown below with $GT_i$ and $P_i$ representing the ground truth and predicted sets of pixels in matched set i.
\begin{equation}
 \text{SEG} = \sum_i\frac{GT_i \cap P_i}{GT_i \cup P_i}
 \label{eq:seg_metric}
\end{equation}
\textit{DET}: Detection Accuracy Measure (DET) \cite{matula_cell_2015} quantifies a model’s instance level detection performance without considering exact segmentation boundaries. This is done by computing the Acyclic Oriented Graph Matching Measure for Detection (AOGM-D) in which an independent graph is generated for the ground truth and model predictions with each cell serving as a node. Then, the number of graph operations (e.g., add node, remove node, split node) that are required to convert the predicted graph into the ground-truth graph is calculated. This value is normalized by the number of graph operations needed to convert an empty graph into the ground truth graph, denoted $AOGM-D_0$. The final metric is computed as follows \ref{eq:det_metric}:
\begin{equation}
 \text{DET} = 1 - \frac{min(\text{AOGM-D}, \text{AOGM-D}_0)}{\text{AOGM-D}_0}
 \label{eq:det_metric}
\end{equation}
\textit{MMA}: Maximum Matching Accuracy (MMA) is a new metric that measures the total intersection area between ground-truth instance masks and an optimally matched set of predicted masks, normalized by the union of all ground-truth and predicted mask areas. To determine the optimal matching, MMA formulates a maximum bipartite matching problem in which ground truth and predicted instances form two disjoint node sets. Edges are constructed between overlapping ground-truth and predicted masks and are weighted by their intersection area.  The resulting matching maximizes total overlap under a one-to-one constraint, ensuring that each predicted mask is matched to at most one ground-truth mask and vice versa. This is shown below \ref{eq:mma_metric} where $GT_i$ and $P_i$ represent the ground truth and predicted sets of pixels in matched set i and GT and P represent the full set of pixels in the ground truth and predicted sets.  
\begin{equation}
 \text{MMA} = \frac{\sum_i GT_i \cap P_i}{GT \cup P}
 \label{eq:mma_metric}
\end{equation}

\subsubsection{Inter-Model Comparisons}
We directly evaluate DINOCell against two of the leading publicly-released cell segmentation models: SAMCell and Cellpose-SAM. Comparison is done on both the test set of a common training source (LIVECell) and three zero-shot dataset containing images that are out of the distribution of each model’s training set. In this way, we compare each model’s ability to fit to a dataset and generalize to unseen images. SEG, DET, and MMA are the chosen metrics for all model comparisons. We utilize ignore masking during evaluation of all models in order to ensure fair comparison between them that minimizes bias from annotation error.

\section{Experiments and Results}
\subsection{Results from Each Training Stage}
As DINOCell progresses through our three training stages, its ViT \cite{dosovitskiy_image_2021} encoder becomes increasingly organized with respect to cellular morphology. By visualizing the first three principal components \cite{jolliffe_pca_2002} between the image encoder patch embeddings, we gain insight into the internal structure of the encoder’s representations. To create this visualization, we input a common image from the cpg0024 subset of the CPG \cite{weisbart_cell_2024} dataset into an encoder from each training stage and retrieve the patch embeddings that they produce. We find the first three principal components of each set of embeddings, project the patch embeddings onto these components, standardize them, and reshape the projected embeddings into a spatial grid such that the three principal directions represent red, green, and blue intensities. 

\begin{figure}[htbp]
\centering
\includegraphics[width=1\textwidth]{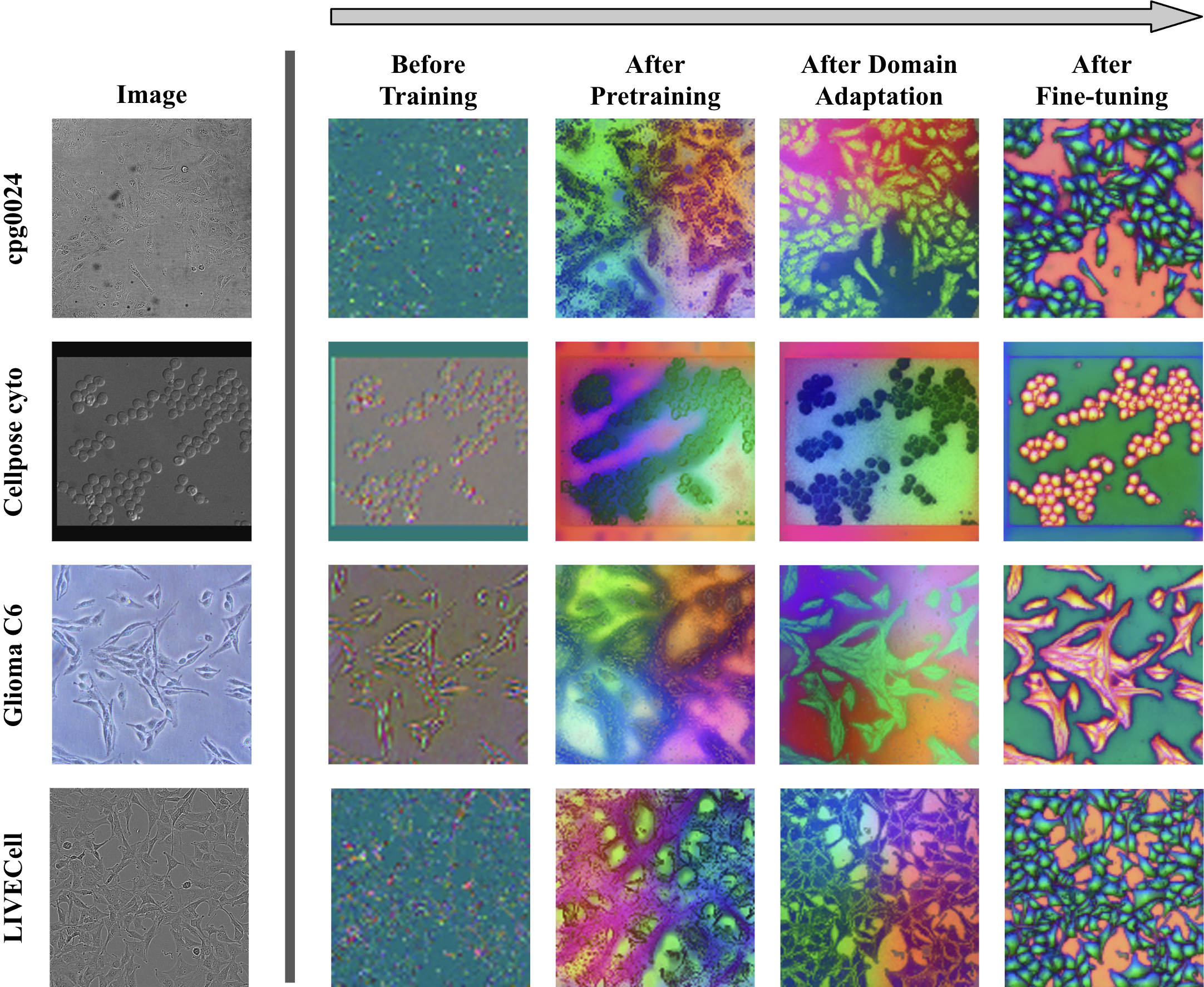}
\caption{\textbf{Evolution of Image Encoder Representations through DINOCell Training.} We compute a PCA between image encoder patches embeddings and show their first three components where each component is mapped to a different color channel (red, green, blue). This is done for four images from different datasets (LIVECell \cite{edlund_livecelllarge-scale_2021}, Glioma C6 \cite{malashin_glioma_2025}, Cellpose cyto \cite{stringer_cellpose_2021}, and cpg0024 \cite{weisbart_cell_2024}) and at each stage of DINOCell’s three stage training process. Clear, per-object structure emerges showing an evolutionary path from initialization to fine-tuning.}
\label{fig:Encoder_Evolution}
\end{figure}

From Figure \ref{fig:Encoder_Evolution}, the PCA visualizations reveal a progressive change in the organization of patch embeddings across the three training stages. With each subsequent stage, overall noise is reduced in the PCA images as encodings begin to cluster into objects. At initialization, the principal component maps appear relatively noisy and exhibit limited correspondence with individual cell boundaries. Using pretrained DINOv2 \cite{oquab_dinov2_2024} weights, the encoder begins to roughly separate foreground and background and detect object edges. After domain adaptation, coherent regions begin to emerge that align more closely with cell morphology. Following supervised fine-tuning, these regions become sharper and often correspond to individual cell interiors and borders. These visualizations provide qualitative evidence that encoder features become more aligned with cellular structures as training progresses.

\subsection{Effect of DINOv2 Pretraining}
We examine the specific effects that both DINOv2 pretraining and domain adaptation have on the representational capacity and performance of DINOCell. Four models are compared: 1) DINOCell without any pretraining, 2) DINOCell with pretraining on only the domain adaptation dataset, 3) DINOCell initialized with DINOv2 weights but not domain adapted, and 4) DINOCell initialized with DINOv2 weights and domain adapted to the microscopy domain. Performance is evaluated on both the fine-tuning testing set and the zero-shot datasets. 

\begin{figure}[htbp]
\centering
\includegraphics[width=1\textwidth]{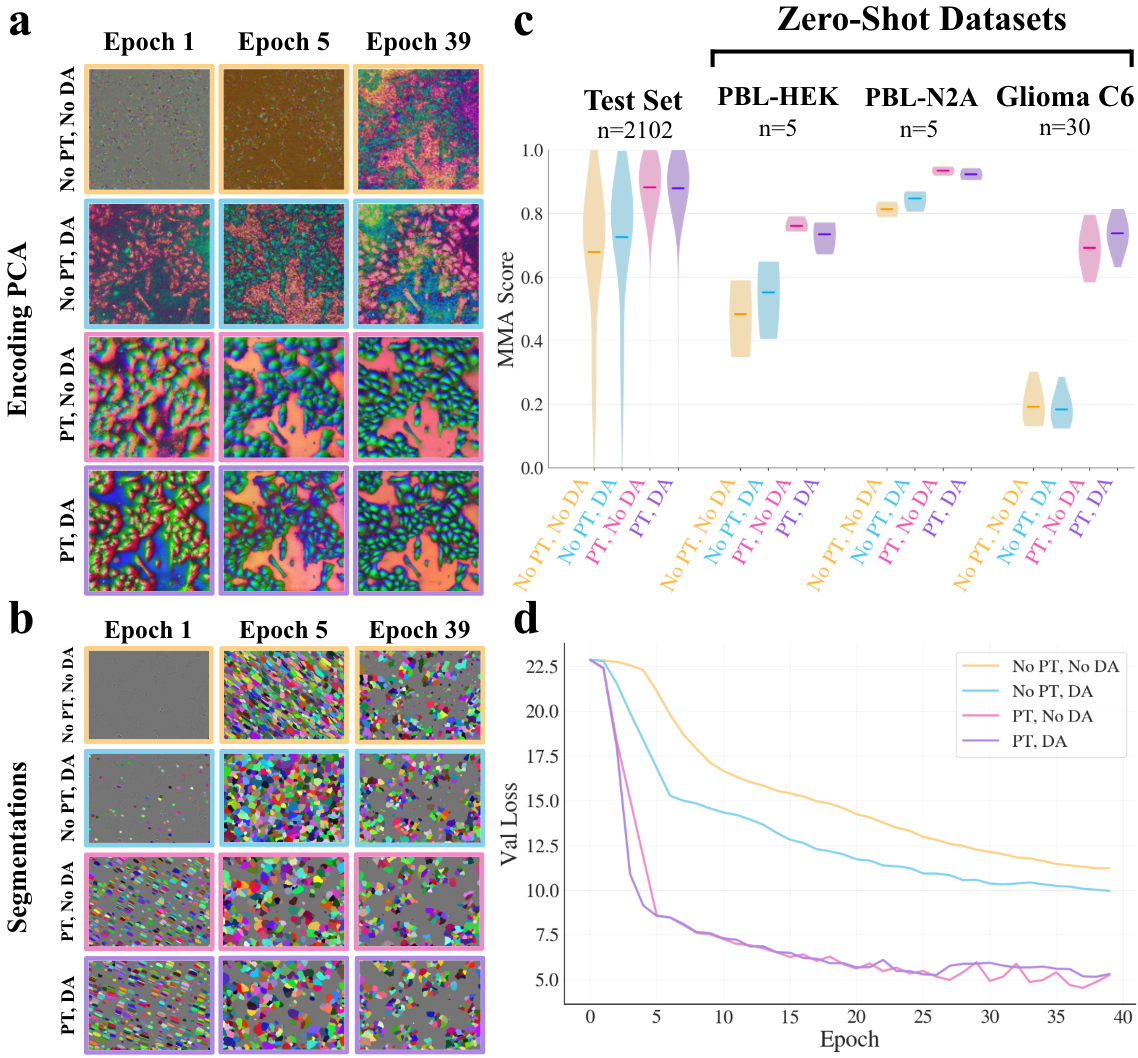}
\caption{\textbf{Effects of pretraining (PT) and domain adaptation (DA) on model performance and encoder representations.} Across epochs, we find that there is a significant effect of self-supervised pretraining on both model performance and encoder representations, while effects of domain adaptation are less significant. (a) Encoder representations of a single image from the cpg0024 \cite{weisbart_cell_2024} subset of the Cellpainting Gallery \cite{weisbart_cell_2024} across epochs using different combinations of PT and DA show that PT yields highly structured representations after 5+ epochs, making segmentation ostensibly appear trivial. (b) Segmentation of cells from a single image of the LIVECell \cite{edlund_livecelllarge-scale_2021} test set across epochs using different combinations of PT and DA show that PT yields highly accurate segmentations and much faster model convergence compared to no-PT. (c) MMA scores across both test and zero-shot datasets show that PT, with and without DA, enhances both in-distribution performance and out-of-distribution generalization. (d) During training, validation loss curves reveal that PT converges much faster than no-PT.}
\label{fig:Pretraining_Effects}
\end{figure}

The results of these pretraining experiments (displayed in Figure \ref{fig:Pretraining_Effects}) suggest a strong positive impact of pretraining on model performance. Specifically, in comparison to a model without any pretraining, the pretrained and domain-adapted DINOCell model increases MMA scores on the fine-tuning test set by 29.51\% and on the zero-shot datasets (PBL-HEK, PBL-N2A, and Glioma C6) by 52.29\%, 36.02\%, and 285.49\%, respectively. In addition, the pretrained model output segmentations are visibly more precise and less prone to both clumping dense cells (combining multiple cells into a single segmentation mask) and segmenting image artifacts. PCA of model embeddings reveals that pretraining enables the development of significantly more refined representational features during fine-tuning. Visualization of the first three of these principal components after fine-tuning shows clear separation of individual objects in pretrained models and highly noisy structures in randomly initialized models. A final key improvement observed in pretrained models is the much faster rate at which they converge and the relatively low number of epochs that they required to produce interpretable segmentations compared to models that were not pretrained. 

Notably, domain adaptation had a much smaller impact on model performance after fine-tuning. MMA scores between pretrained models that did and did not undergo domain adaptation were nearly the same on the fine-tuning test set and the PBL-HEK and PBL-N2A zero-shot sets. A modest improvement of 6.48\% is obtained by the domain-adapted model on the Glioma C6 zero-shot dataset, potentially indicating a positive benefit of domain adaptation for model generalization to out-of-domain images but further experiments would need to be conducted to state this more concretely. Despite improvements in the clarity of encoder PCA visualizations observed during domain adaptation, the domain-adapted and non-domain-adapted pretrained models converge to virtually identical representations. No significant strengths or weaknesses can be visibly observed between domain-adapted and non-domain-adapted models in the precision of their segmentations either. Domain-adapted models do produce a small boost in performance during very early training epochs but they quickly converge to similar loss levels as non-domain-adapted models. 

Lastly, we observe that pretraining on unlabeled cell images from random initialization produces meaningful performance boosts during fine-tuning compared to models that undergo no pretraining. The model that pretrained on only the Domain Adaptation Dataset achieved MMA scores that were 6.84\% higher on the fine-tuning test set and 14.44\% and 4.10\% higher on the PBL-HEK and PBL-N2A zero-shot datasets but 4.68\% lower on the Glioma C6 zero-shot dataset compared to the model that began fine-tuning from random initialization. Additionally, this lightly-pretrained model produces clearer image encodings and more accurate segmentation masks that appear far earlier in fine-tuning than the randomly initialized model’s outputs. However, the performance of this lightly-pretrained model is far closer to that of the randomly initialized model than it is to the performance of the models that use DINOv2 pretrained weights.

\subsection{Performance on Individual Datasets}
\begin{figure}[htbp]
\centering
\includegraphics[width=1\textwidth]{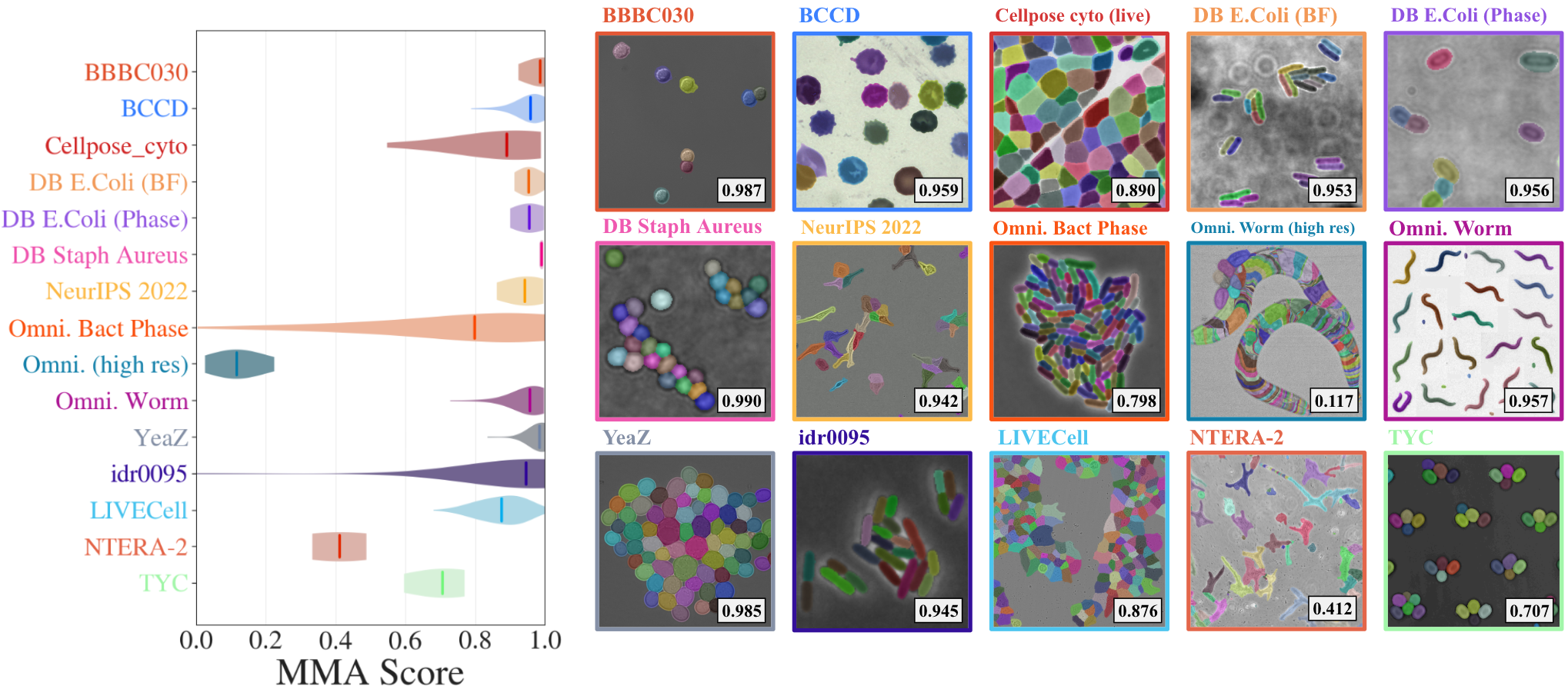}
\caption{\textbf{DINOCell per-dataset performance on the test split of the Finetuning Dataset, shown for both (a) aggregate performance, and (b) representative images.} \textbf{(a)} Maximum Matching Accuracy (MMA) on the test split of each dataset used for fine-tuning. MMA evaluates how well predicted instances correspond to ground-truth cells (maximum score = 1.0). DINOCell achieves high performance on most datasets, indicating strong detection and segmentation accuracy across a range of cell morphologies and imaging conditions. \textbf{(b)} Representative images from each dataset with predicted segmentations overlaid. These qualitative examples illustrate DINOCell’s ability to accurately delineate cells with diverse shapes, densities, and imaging modalities.}
\label{fig:Individual_Dataset_Performance}
\end{figure}

Evaluating DINOCell on the testing test of each dataset individually reveals the profound strengths of this model. From Figure \ref{fig:Individual_Dataset_Performance}, we observe high accuracy (0.94+ MMA) on cells with simple shapes and clear outlines (DeepBacs [DB] Staph Aureus, DeepBacs Ecoli [BF], DeepBacs E.Coli [Phase], YeaZ, BBBC030, BCCD, Omnipose [Omni.] Worm, and idr0095) and even on some datasets with cells of more diverse morphologies (NeurIPS\_2022). On images containing cells with significantly more complex morphology (LIVECell, Cellpose cyto) or high density cultures (Omnipose Bact Phase), performance dropped only slightly (~0.8-0.9 MMA) suggesting robustness to diverse cell types and lighting conditions. 

DINOCell performance dipped to low levels  (< 0.71 MMA) on only three datasets: TYC, NTERA-2, and Omnipose Worm (high res). In the TYC dataset, DINOCell accurately segments the microstructures in the dataset but frequently assigns separate labels to disjoint components that correspond to a single ground-truth instance. As a result, predicted segmentations often contain multiple objects where the dataset annotation defines only one. In the NTERA-2 dataset, which contains preneuronal embryonic cells with extended axonal connections, DINOCell often segments the thin axonal structures into multiple fragments rather than maintaining a continuous instance across the entire structure. Finally, DINOCell achieves its lowest performance on the Omnipose Worm (high res) dataset. These images contain large, elongated C. elegans worms occupying a majority of the image. In many cases, predicted segmentations fragment the worm into many instances rather than producing a single continuous mask. 

\subsection{Model Comparison}
\subsubsection{Other Models}
We compare DINOCell against two leading models in cell segmentation: SAMCell and Cellpose-SAM.

\textit{SAMCell}: SAMCell is a ViT-B model initialized with SAM weights that attempts to predict a distance map from an input image and uses watershed to decode segmentation masks. Two separate models are trained (SAMCell-LIVECell and SAMCell-cyto), each on only one dataset (LIVECell and Cellpose-cyto), making them specialist models. We evaluate LIVECell test set accuracy on the LIVECell pretrained model and zero-shot accuracy on both of the models. Inference on this model is done using the test.ipynb notebook in the SAMCell repository with the trained model weights they provide in the same repository.

\textit{Cellpose-SAM}: Cellpose-SAM is a ViT-L generalist cell segmentation model initialized with SAM weights and trained on a similar dataset as DINOCell with the addition of fluorescence and tissue images but without the BBBC030, idr0095, NTERA-2, or TYC datasets. DINOCell uses the same flows objective and gradient-tracking algorithm for segmentation mask decoding as were used for the Cellpose-SAM model. Unlike DINOCell or SAMCell, this model is trained and evaluated on rgb images whereas the other models internally convert the images to grayscale. We conduct inference on Cellpose-SAM using the run\_Cellpose-SAM.ipynb notebook from their repository which loads in their pretrained model. 

\subsubsection{LIVECell Test Set Comparison}
\begin{figure}[htbp]
\centering
\includegraphics[width=0.9\textwidth]{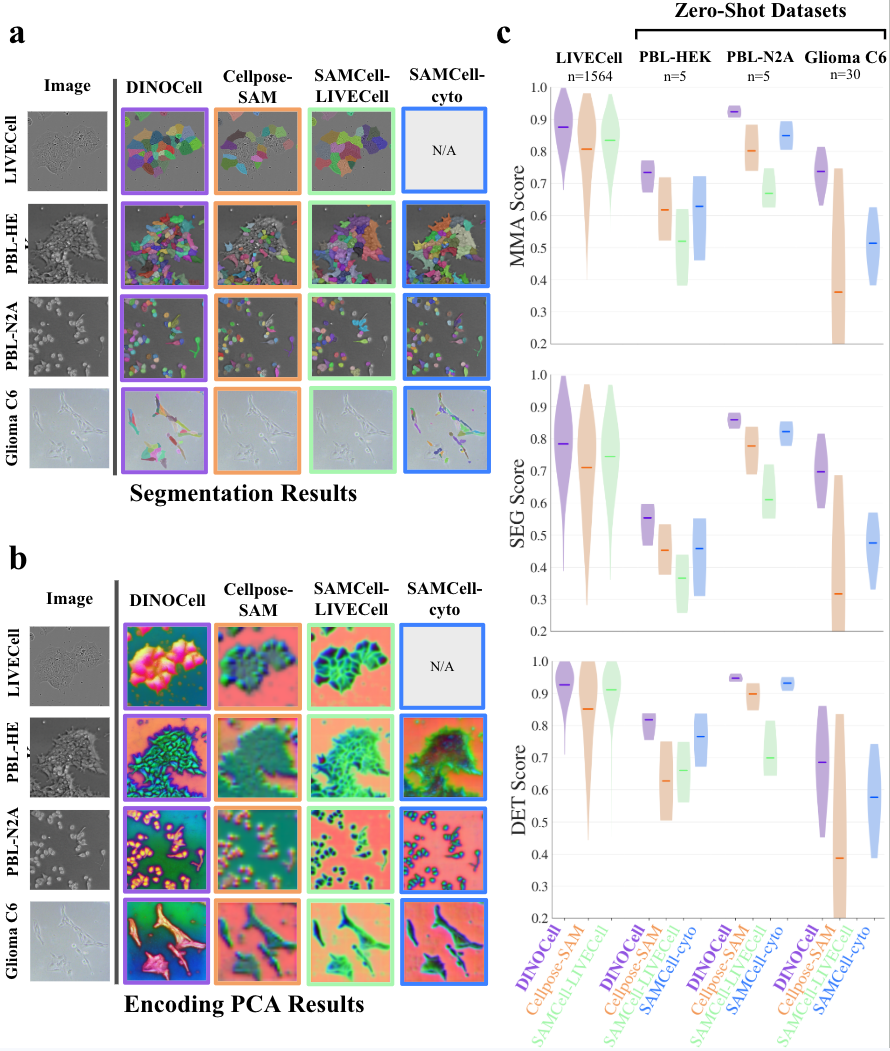}
\caption{\textbf{DINOCell outperforms Cellpose-SAM , SAMCell-LIVECell, and SAMCell-cyto both visually and quantitatively across all metrics.} \textbf{(a)} Qualitative results are shown using segmentation outputs for representative images across the LIVECell \cite{edlund_livecelllarge-scale_2021} and zero-shot datasets.  On all data sets, the cells appear visually to be thoroughly and accurately segmented as compared to models from prior publications on the same images.  Specifically, cells are not missed or clumped and artifacts are absent, relatively speaking. \textbf{(b)} Intermediate Principal Component Analysis (PCA) of the encoder output.  PCA of DINOCell encodings reveal per-object features with well-defined boundaries that do not emerge as clearly or completely in other models. \textbf{(c)} Maximum Matching Accuracy (MMA), Segmentation Accuracy Metric (SEG), and Detection Accuracy Metric (DET) shown for DINOCell, Cellpose-SAM, and SAMCell-LIVECell model outputs on the LIVECell testing. Zero-shot performance is also shown for all prior models with the addition of the SAMCell-cyto model. DINOCell displays dominance with the highest MMA, SEG, and DET scores across all datasets.}
\label{fig:Model_Comparison}
\end{figure}

We first compare the three models on the test split of the LIVECell dataset, whose training set was included in each model’s fine-tuning. From Figure \ref{fig:Model_Comparison}(c), we observe that DINOCell surpasses baseline performance across all metrics, beating even the specialist SAMCell-LIVECell model. Specifically, DINOCell achieves MMA, SEG, and DET scores of 0.875, 0.784, and 0.926, which correspond to 4.99\%, 5.26\%, and 1.67\% improvements relative the the specialist model SAMCell-LIVECell and 8.49\%, 10.42\%, and 8.74\% improvements relative to the generalist model Cellpose-SAM. This indicates DINOCell’s increased capacity for fitting the particular nuances of a single dataset while training on a much larger, more diverse set of images. 
	
Additionally, row 1 of Figure \ref{fig:Model_Comparison}(a) qualitatively displays the heightened accuracy of DINOCell segmentations relative to the competitor models.  Cellpose-SAM significantly under-segments the image, missing a large number of cells. SAMCell-LIVECell accurately identifies all parts of the image containing cells, but tends to clump many cells under one segmentation mask. In contrast, the DINOCell predictions are both precise and thorough, identifying each individual cell and sharply defining its boundaries. DINOCell demonstrates far greater resistance to these common failure cases of under-segmentation and cell clumping. 

\subsubsection{Zero-Shot Comparison}
The most significant contribution of DINOCell is its ability to generalize to out-of-distribution images. In Figure \ref{fig:Model_Comparison}(c), we can see that DINOCell’s performance in zero-shot evaluation nearly rivals that of its performance on the testing set and well exceeds the baseline models. The SAMCell-cyto model achieves the next highest scores across all datasets and metrics.

For all models, zero-shot performance is highest on the PBL-N2A dataset composed of images containing Neuro-2a (N2A) cells. From row 3 of Figure \ref{fig:Model_Comparison}(a), we see that the majority of these cells are highly uniform with simple, circular shapes, little clumping, and clear boundaries. DINOCell achieves MMA, SEG, and DET scores of 0.923, 0.859, and 0.947 on this dataset, surpassing SAMCell-cyto by 8.73\%, 4.61\%, and 1.59\% respectively. In contrast, the Human Embryonic Kidney (HEK) cells from the PBL-HEK dataset have more complex and highly variable morphologies. The HEK cells are often clustered into dense clumps, and are imaged on a wide range of magnification levels. These conditions increase the difficulty of segmentation and thus lower the metric scores. On the HEK dataset, DINOCell achieves MMA, SEG, and DET scores of 0.734, 0.553, and 0.818, improving on SAMCell-cyto by 16.84\%, 20.71\%, and 6.89\% respectively.

We observe the largest disparity between DINOCell and competitor models on the Glioma C6 dataset. As mentioned earlier, this dataset is built for robustness evaluation and contains highly variable lighting conditions, magnification levels, and cell morphologies. All baseline models greatly struggle on this dataset: SAMCell-LIVECell does not segment a single cell in the entire dataset, SAMCell-cyto misses many cells and produces jagged and incomplete segmentations, and Cellpose-SAM fails to segment more than half of the cells in many images and doesn’t segment any cells at all in others. DINOCell, however, segments these images with high accuracy, including MMA, SEG, and DET scores of 0.737, 0.697, and 0.684, surpassing the leading baseline (SAMCell-cyto) scores by 43.84\%, 46.84\%, and 18.81\%. This result further validates DINOCell's robustness on highly variable microscopy images and the efficacy of self-supervised pretraining and domain adaptation.

\subsubsection{Image Encoder PCA Comparison}
For a final comparison between these models, we qualitatively examine the underlying representational structures in each model’s image encoder. DINOCell, SAMCell, and Cellpose-SAM are all encoder-heavy models that utilize only lightweight decoding architectures. Therefore, by investigating their encoder representations we gain insight into their structural and object-level priors. To conduct this comparison, we visualize the first three principal components of the encoder representations of images from both the LIVECell test set and each zero-shot dataset 

In Figure \ref{fig:Model_Comparison}(b), we observe interpretable structure in each of the model’s encodings. However, the sharpness and clarity of object-level separation in the principal component visualizations of DINOCell encodings far exceeds that found in the other models. In particular, the PCA visualization on the PBL-HEK image in the second row of Figure \ref{fig:Model_Comparison}(a) demonstrates precise object identification and partitioning of a highly dense cell cluster that is not emulated in other models’ representations. 

\section{Conclusion}
In this work, we introduced DINOCell, a cell segmentation framework that leverages self-supervised visual representations to improve segmentation performance in microscopy images. By adapting a pretrained DINOv2 vision transformer to the microscopy domain prior to supervised fine-tuning, the model learns well-structured features aligned with cellular morphology. Experiments demonstrate consistent improvements from DINOv2 pretraining including MMA increases of 29.51\% and 36-285\% on test and zero-shot evaluations compared to non-pretrained models. Notably, our results show that large-scale self-supervised pretraining provides strong performance gains even without explicit domain adaptation. Compared to leading segmentation-pretrained baseline models (SAMCell and Cellpose-SAM), DINOCell increases MMA scores by at least 5\% on the LIVECell test set and 8-43\% in zero-shot evaluation. Qualitatively, visualization of the principal components of each model's image encoder reveals highly improved boundaries and object separation in DINOCell encodings. This improvement in representation quality translates directly into better segmentation boundaries, fewer missed cells, and decreased clumping of cells. Overall, these findings highlight the effectiveness of self-supervised representation learning for tasks operating on data distributions that differ substantially from those used to train large-scale foundation models.

\section*{Acknowledgements}
CRF acknowledges the NIH R01NS102727, NIH Single Cell Grant 1 R01 EY023173, NIH R01DA029639 and NIH RF1AG079269, support from Georgia Tech through the Institute for Bioengineering and Biosciences, Invention Studio, and the George W. Woodruff School of Mechanical Engineering.

\bibliographystyle{unsrt}
\bibliography{DINOCell}  %%% Uncomment this line and comment out the ``thebibliography'' section below to use the external .bib file (using bibtex) .

\section{Supplementary Materials}
\subsection{Fine-tuning Dataset Descriptions}
\textit{BBBC030 \cite{koos_dic_2016}}: The BBBC030 dataset contains 60 images of Chinese Hamster Ovary (CHO) cells. The images were taken on a DIC microscope with a 20x magnification.  The dataset is available here: https://bbbc.broadinstitute.org/BBBC030. 
\newline
\newline
\textit{BCCD \cite{dpto_blood_2023}}: The BCCD dataset contains 1,169 train and 159 test images of stained blood cells from a blood smear. The dataset is available here: https://www.kaggle.com/datasets/jeetblahiri/bccd-dataset-with-mask/data. 
\newline
\newline
\textit{Cellpose cyto \cite{stringer_cellpose_2021}}: The Cellpose cyto dataset contains 796 train and 68 test images from varying modalities and even includes some non-biological dataset. We take only the images that are DIC, PC, or BF, labeled by whole cell, and of biological cells, leaving 180 train and 20 test images. The dataset is available here: https://www.cellpose.org/dataset. 
\newline
\newline
\textit{DeepBacs E.Coli (BF) \cite{spahn_deepbacs_ecoli_bf_oct_2021}}: The DeepBacs E.Coli (BF) dataset contains 20 train and 16 test images of live E.Coli cells. The images were taken on a BF microscope with a 100x magnification. The dataset is available here:https://zenodo.org/records/5550935.  
\newline
\newline
\textit{DeepBacs E.Coli (Phase) \cite{spahn_deepbacs_2022}}: The DeepBacs E.Coli (Phase) dataset contains 34 train and 13 test images of live E.Coli cells after release from stationary phase. The images were taken on a BF microscope with a 100x magnification. The dataset is available here: https://zenodo.org/records/6400327.
\newline
\newline
\textit{DeepBacs Staph Aureus \cite{pereira_deepbacs_2021}}: The DeepBacs Staph Aureus dataset contains 8 train and 6 test images of live S.Aureus cells. The images were taken on a BF microscope with a 60x magnification. The dataset is available here: https://zenodo.org/records/5550933. 
\newline
\newline
\textit{NeurIPS 2022 Challenge \cite{ma_neurips_2024}}: The NeurIPS 2022 Challenge dataset contains 1000 train and 50 test images of diverse cells from varying origins and morphologies. We take only the images that are DIC, PC, or BF, labeled by whole cell, leaving 471 train and 21 test images. The dataset is available here: https://neurips22-cellseg.grand-challenge.org/dataset/.  
\newline
\newline
\textit{Omnipose Bact Phase \cite{cutler_omnipose_2022}}: The Omnipose Bact Phase dataset contains 249 train and 148 test images of bacteria. The images were taken on a phase-contrast microscope. The dataset is available here: https://osf.io/xmury/overview. 
\newline
\newline
\textit{Omnipose Worm (high res) \cite{cutler_omnipose_2022}}: The Omnipose Worm (high res) dataset contains 48 train and 11 test images of C.elegans. The images were taken on a phase-contrast microscope with a 15x magnification. The dataset is available here: https://osf.io/xmury/overview. 
\newline
\newline
\textit{Omnipose Worm \cite{cutler_omnipose_2022}}: The Omnipose Worm dataset contains 60 train and 60 test images of C.elegans. The images were taken on a phase-contrast microscope. The dataset is available here: https://osf.io/xmury/overview. 
\newline
\newline
\textit{YeaZ \cite{dietler_convolutional_2020}}: The YeaZ dataset contains 349 images of yeast cells. The images were taken in both brightfield (306) and phase-contrast (43). The dataset is available here: https://www.epfl.ch/labs/lpbs/data-and-software/.
\newline
\newline
\textit{idr0095 \cite{williams_image_2017}}: The idr0095 dataset contains 347 train and 87 test images of E.Coli cells. The images were taken on a phase-contrast microscope. The dataset is available here: https://idr.openmicroscopy.org/webclient/?show=project-1401.
\newline
\newline
\textit{LIVECell \cite{edlund_livecelllarge-scale_2021}}: The LIVECell dataset contains 3823 train and 1564 test images of eight different cell types (A172, BT474, BV2, Huh7, MCF7, SHSY5Y, SkBr3, SKOV3). The images were taken on a phase-contrast microscope. The dataset is available here: https://sartorius-research.github.io/LIVECell/.
\newline
\newline
\textit{NTERA-2 \cite{zhang_morphoseg_2025}}: The NTERA-2 dataset contains 30 train and 8 test images of Ntera-2 cells. The images were taken on a brightfield microscope with 10x and 20x objectives. The dataset is available here: https://orda.shef.ac.uk/articles/dataset/\_b\_Dataset\_of\_Ntera-2\_cells\_for\_Segmentation\_b\_/25604421?file=48999580.
\newline
\newline
\textit{TYC \cite{vijayan_deep_2024}}: The TYC dataset contains 82 train and 9 test images of yeast cells in microstructures. The images were taken on a brightfield microscope. The dataset is available here: https://tudatalib.ulb.tu-darmstadt.de/items/b6c09b46-bd02-45f9-a401-e60c078e65f4.

\subsection{Code Availability}
The DINOCell code is available at https://github.com/kadenstillwagon/DINOCell. The highest performance DINOCell model (used for all comparisons in the "Model Comparison" section) has its weights released under the "models" folder. Evaluation can be conducted using the "demo.ipynb" notebook in the "src" folder.

\end{document}